# OBDD-based Universal Planning for Synchronized Agents in Non-Deterministic Domains

**Rune M. Jensen**                                    RUNEJ@CS.CMU.EDU
**Manuela M. Veloso**                                 MMV@CS.CMU.EDU
*Computer Science Department, Carnegie Mellon University*
*Pittsburgh, PA 15213-3891, USA*

## Abstract

Recently model checking representation and search techniques were shown to be efficiently applicable to planning, in particular to non-deterministic planning. Such planning approaches use Ordered Binary Decision Diagrams (OBDDs) to encode a planning domain as a non-deterministic finite automaton and then apply fast algorithms from model checking to search for a solution. OBDDs can effectively scale and can provide universal plans for complex planning domains. We are particularly interested in addressing the complexities arising in non-deterministic, multi-agent domains. In this article, we present UMOP, a new universal OBDD-based planning framework for non-deterministic, multi-agent domains. We introduce a new planning domain description language, *NADL*, to specify non-deterministic, multi-agent domains. The language contributes the explicit definition of controllable agents and uncontrollable environment agents. We describe the syntax and semantics of *NADL* and show how to build an efficient OBDD-based representation of an *NADL* description. The UMOP planning system uses *NADL* and different OBDD-based universal planning algorithms. It includes the previously developed strong and strong cyclic planning algorithms. In addition, we introduce our new optimistic planning algorithm that relaxes optimality guarantees and generates plausible universal plans in some domains where no strong nor strong cyclic solution exists. We present empirical results applying UMOP to domains ranging from deterministic and single-agent with no environment actions to non-deterministic and multi-agent with complex environment actions. UMOP is shown to be a rich and efficient planning system.

## 1. Introduction

Classical planning involves the automatic generation of actions to traverse a state space to achieve specific goal states. Various algorithms have been developed to address the state-action representation and the search for actions. Traditionally these algorithms have been classified according to their search space representation as either state-space planners (e.g., PRODIGY, Veloso et al., 1995) or plan-space planners (e.g., UCPOP, Penberthy & Weld, 1992).

A new research trend has been to develop new encodings of planning problems in order to adopt efficient algorithms from other research areas, leading to significant developments in planning algorithms, as surveyed by Weld (1999). This class of planning algorithms includes GRAPHPLAN (Blum & Furst, 1997) that uses a flow-graph encoding to constrain the search and SATPLAN (Kautz & Selman, 1996) that encodes the planning problem as a satisfiability problem and uses fast model satisfaction algorithms to find a solution.

Recently, another new planner MBP (Cimatti et al., 1997) was introduced that encodes a planning domain as a non-deterministic finite automaton (NFA) represented by an Ordered Binary Decision Diagram (OBDD, Bryant, 1986). In contrast to the previous algorithms, MBP effectively extends to non-deterministic domains producing universal plans as robust solutions. Due to the scalability of the underlying model checking representation and search techniques, it can be shown to be a very efficient non-deterministic planner (Cimatti et al., 1998a, 1998b).

One of our main research objectives is to develop planning systems suitable for planning in uncertain, single, or multi-agent environments (Haigh & Veloso, 1998; Veloso et al., 1998; Stone & Veloso, 1998). The universal planning approach, as originally developed (Schoppers, 1987), is appealing for this type of environments. A universal plan is a set of state-action rules that aim at covering the possible multiple situations in the non-deterministic environment. A universal plan is executed by interleaving the selection of an action in the plan and observing the resulting effects in the world. Universal planning resembles the outcome of reinforcement learning (Sutton & Barto, 1998), in that the state-action model captures the uncertainty of the world. Universal planning is a precursor approach,[1] where all planning is done prior to execution, building upon the assumption that a non-deterministic model of the execution environment can be acquired, and leading therefore to a sound and complete planning approach.

However, universal planning has been criticized (e.g., Ginsberg, 1989), due to a potential exponential growth of the universal plan size with the number of propositions defining a domain state. An important contribution of MBP is thus the use of OBDDs to represent universal plans. In the worst case, this representation may also grow exponentially with the number of domain propositions, but because OBDDs are very compact representations of boolean functions, this is often not the case for domains with a regular structure (Cimatti et al., 1998a). Therefore, OBDD-based planning seems to be a promising approach to universal planning.

MBP specifies a planning domain in the action description language $\mathcal{AR}$ (Giunchiglia et al., 1997) and translates it to a corresponding NFA, hence limited to planning problems with finite state spaces. The transition relation of the automaton is encoded as an OBDD that allows for the use of efficient breadth-first search techniques developed for model checking (McMillan, 1993). MBP includes two algorithms for universal planning. The *strong planning* algorithm tries to generate a plan that is guaranteed to achieve the goal for all of the possible outcomes of the non-deterministic actions. If no such strong solution exists, the algorithm fails. The *strong cyclic planning* algorithm returns a strong solution, if one exists, or otherwise tries to generate a plan that may contain loops but is guaranteed to achieve the goal, given that all cyclic executions eventually terminate. If no such strong cyclic solution exists, the strong cyclic planning algorithm fails.

In this article we present our OBDD-based planning system, UMOP (Universal Multi-agent Obdd-based Planner), that uses a new OBDD-based encoding, generates universal plans in multi-agent non-deterministic domains, and includes a new "optimistic" planning algorithm.

---

1. The term *precursor* originates in Dean et al. (1995) in contrast to *recurrent* approaches that replan to recover from execution failures.

Our overall approach for designing an OBDD-based planner is similar to the approach developed for MBP. Our main contribution is an efficient encoding of a new front end domain description language, *NADL* (Non-deterministic Agent Domain Language). *NADL* has more resemblance with previous planning languages than the action description language $\mathcal{AR}$ currently used by MBP. It has powerful action descriptions that can perform arithmetic operations on numerical domain variables. Domains comprised of synchronized agents can be modelled by introducing concurrent actions based on a multi-agent decomposition of the domain.

In addition, *NADL* introduces a separate and explicit environment model defined as a set of *uncontrollable* agents, i.e., agents whose actions cannot be a part of the generated plan. *NADL* has been carefully designed to allow for efficient OBDD-encoding. Thus, UMOP contributes a partitioned transition relation representation of the NFA that is known from model checking to scale up well (Burch et al., 1991; Ranjan et al., 1995). Our empirical experiments suggest that this is also the case for UMOP.

UMOP includes the previously developed algorithms for OBDD-based universal planning. In addition, we introduce a new "optimistic" planning algorithm that relaxes optimality guarantees and generates plausible universal plans in domains where no strong nor strong cyclic solution exists.

The article is organized as follows. Section 2 discusses previous approaches to planning in non-deterministic domains. Section 3 gives a brief overview of OBDDs and NFA encodings. It may be skipped by readers already familiar with the subject. Section 4 introduces *NADL*, shows how to encode a planning problem, and formally describes the syntax and semantics of this description language in terms of an NFA. We also discuss the properties of the language based on an example and argue for our design choices. Section 5 presents the OBDD representation of *NADL* domain descriptions. Section 6 describes the different algorithms that have been used for OBDD-based planning and introduces our optimistic planning algorithm. Section 7 presents empirical results in several planning domains, ranging from single-agent and deterministic ones to multi-agent and non-deterministic ones. We experiment with previously used domains and introduce two new ones, namely a power plant and a soccer domain, as non-deterministic, multi-agent planning problems. Finally, Section 8 draws conclusions and discusses directions for future work.

## 2. Related Work

Recurrent approaches performing planning interleaved or in parallel with execution have been widely used in non-deterministic robotic domains (e.g., Georgeff & Lansky, 1987; Gat, 1992; Wilkins et al., 1994; Haigh & Veloso, 1998). A group of planners suitable for recurrent planning is action selectors based on heuristic search (Koenig & Simmons, 1995; Bonet et al., 1997). The min-max LRTA* algorithm (Koenig & Simmons, 1995; Smirnov et al., 1996) can generate suboptimal plans in non-deterministic domains through a search and execution iteration. The search is based on a heuristic goal distance function that must be provided for a specific problem. The ASP algorithm (Bonet et al., 1997) uses a similar approach and further defines a heuristic function for STRIPS-like (Fikes & Nilsson, 1971) action representations. In contrast to min-max LRTA*, ASP does not assume a non-deterministic environment, but is robust to non-determinism caused by action perturbations (i.e., that another action than the planned action is chosen with some probability).

In general, recurrent approaches are incomplete because acting on an incomplete plan can make the goal unachievable. Precursor approaches perform all decision making prior to execution and thus may be able to generate complete plans by taking all possible effects of actions into account. However, they rely on a complete model of the world's uncertainty.

The precursor approaches include conditional (Etzioni et al., 1992; Peot & Smith, 1992; Blythe & Veloso, 1997), probabilistic (Drummond & Bresina, 1990; Dean et al., 1995; Blythe, 1998) and universal planning (Schoppers, 1987; Cimatti et al., 1998a, 1998b; Kabanza et al., 1997). For example, the CNLP partial order, conditional planner handles non-determinism by constructing a conditional plan that accounts for each possible situation or contingency that could arise (Peot & Smith, 1992). At execution time it is determined which part of the plan to execute by performing sensing actions that are included in the plan to test for the appropriate conditions.

Probabilistic planners try to maximize the probability of goal satisfaction, given conditional actions with probabilistic effects. Drummond and Bresina (1990) represent plans as a set of Situated Control Rules (SCRs) (Drummond, 1989) mapping situations to actions. The planning algorithm begins by adding SCRs corresponding to the most probable execution path that achieves the goal. It then continues adding SCRs for less probable paths and may end with a complete plan taking all possible paths into account.

Universal plans differ from conditional and probabilistic plans by specifying appropriate actions for every possible state in the domain. Like conditional and probabilistic plans, universal plans require the world to be accessible in order to execute the plan.

Universal planning was introduced by Schoppers (1987) who used decision trees to represent plans. Recent approaches include Kabanza et al. (1997) and Cimatti et al. (1998a, 1998b). Kabanza et al. (1997) represents universal plans also as a set of Situated Control Rules. Their algorithm incrementally adds SCRs to a final plan in a way similar to Drummond and Bresina (1990). The goal is a formula in temporal logic that must hold on any valid sequence of actions.

Reinforcement Learning (RL) (Sutton & Barto, 1998) can also be regarded as universal planning. In RL the goal is represented by a reward function in a Markov Decision Process (MDP) model of the domain. In the precursor version of RL, the MDP is assumed to be known and a control policy maximizing the expected reward is found prior to execution. The policy can either be represented explicitly in a table or implicitly by a function (e.g., a neural network). Because RL is a probabilistic approach, its domain representation is more complex than the domain representation used by a non-deterministic planner. Thus, we may expect non-deterministic planners to be able to handle domains with a larger state space than RL. But RL may produce policies with a higher quality than a universal plan generated by a non-probabilistic, non-deterministic planner. Moreover, in the recurrent version, RL learns the world model during execution and thus does not require a complete world model prior to execution. Though, in theory it needs infinite execution examples to converge to the optimal universal plan.

All previous approaches to universal planning, except Cimatti et al. (1998a, 1998b), use an explicit representation of the universal plan (e.g., SCRs). Thus, in the general case, an exponential size of the plan in the number of propositions defining a domain state must be expected, as argued by Ginsberg (1989).

The compact and implicit representation of universal plans obtained with OBDDs does not necessarily grow exponentially for regularly structured domains as shown by Cimatti et al. (1998a). Further, the OBDD-based representation of the NFA of a non-deterministic domain enables the application of efficient search algorithms from model checking capable of handling very large state spaces.

## 3. Introduction to OBDDs

An Ordered Binary Decision Diagram (Bryant, 1986) is a canonical representation of a boolean function with $n$ linear ordered arguments $x_1, x_2, ..., x_n$.

An OBDD is a rooted, directed acyclic graph with one or two terminal nodes of out-degree zero labeled 1 or 0, and a set of variable nodes $u$ of out-degree two. The two outgoing edges are given by the functions $high(u)$ and $low(u)$. Each variable node is associated with a propositional variable in the boolean function the OBDD represents. The graph is ordered in the sense that all paths in the graph respect the ordering of the variables.

An OBDD representing the function $f(x_1, x_2) = x_1 \wedge x_2$ is shown in Figure 1. Given an assignment of the arguments $x_1$ and $x_2$, the value of $f$ is determined by a path starting at the root node and iteratively following the high edge, if the associated variable is true, and the low edge, if the associated variable is false. The value of $f$ is *True* or *False* if the label of the reached terminal node is 1 or 0, respectively.

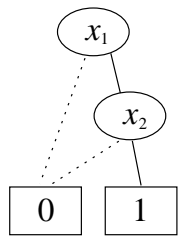

Figure 1: An OBDD representing the function $f(x_1, x_2) = x_1 \wedge x_2$. High and low edges are drawn as solid and dotted lines, respectively.

An OBBD graph is reduced so that no two distinct nodes $u$ and $v$ have the same variable name and low and high successors (Figure 2a), and no variable node $u$ has identical low and high successors (Figure 2b).

The OBDD representation has two major advantages: First, it is an efficient representation of boolean functions because the number of nodes often is much smaller than the number of truth assignments of the variables. The number of nodes can grow exponential with the number of variables, but most commonly encountered functions have a reasonable representation. Second, any operation on two OBDDs, corresponding to a boolean operation on the functions they represent, has a low complexity bounded by the product of their node counts (Bryant, 1986).

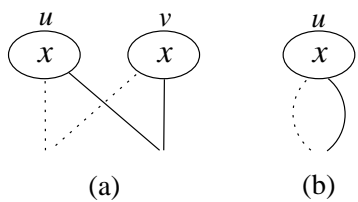

Figure 2: Reductions of OBDDs. (a) nodes associated to the same variable with equal low and high successors will be converted to a single node. (b) nodes causing redundant tests on a variable are eliminated.

A disadvantage of OBDDs is that the size of an OBDD representing some function is very dependent on the ordering of the variables. To find an optimal variable ordering is a co-NP-complete problem in itself, but as illustrated in Figure 3 a good heuristic for choosing an ordering is to locate related variables near each other (Clarke et al., 1999).

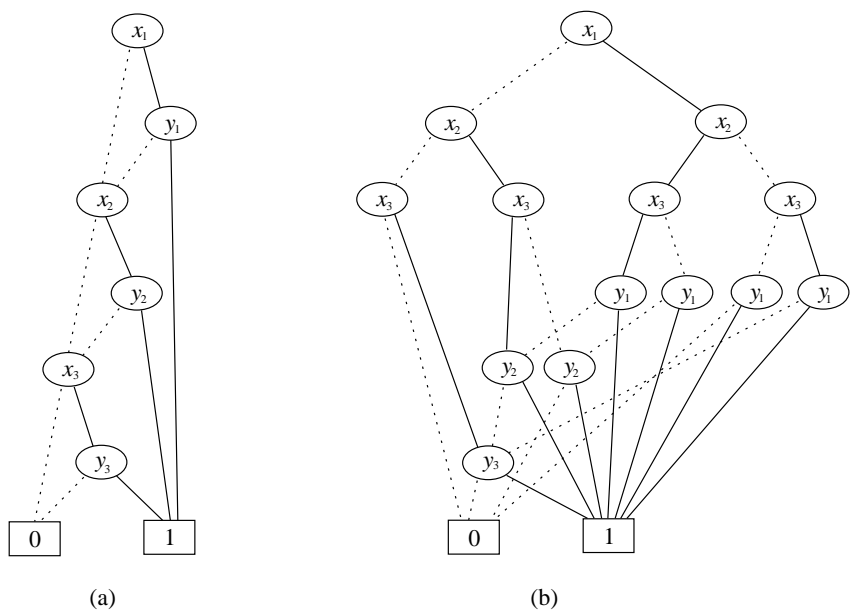

Figure 3: This Figure shows the effect of variable ordering for the expression $(x_1 \wedge y_1) \vee (x_2 \wedge y_2) \vee (x_3 \wedge y_3)$. The OBDD in (a) only grows linearly with the number of variables in the expression, while the OBDD in (b) has an exponential growth. The example illustrates that placing related variables near to each other in the ordering often is a good heuristic.

OBDDs have been successfully applied to model checking. In model checking the behavior of a system is modelled by a finite state automaton with the transition relation represented as an OBDD. Desirable properties are checked by using OBDD manipulations to analyze the state space of the system (Clarke et al., 1986; McMillan, 1993).

Interestingly, a similar approach can be used for solving non-deterministic planning problems. As an example, consider the NFA representation of a non-deterministic planning domain shown in Figure 4a. In this domain there are four states given by the four possible value assignments of the two boolean state variables $x_1$ and $x_2$. Inputs to the NFA denote actions in the domain and are defined by the boolean variable $a$. The OBDD representing the transition relation $T(a, x_1, x_2, x_1', x_2')$ of the NFA is shown in Figure 4b. The definition of $T$ is straightforward: for some assignment of its arguments, $T$ is true iff action $a$ causes a transition from the current state given by the value of $x_1$ and $x_2$ to the next state given by the value of $x_1'$ and $x_2'$.[2] Note that the OBDD representing $T$ for the example turns out not to depend on $x_2'$.

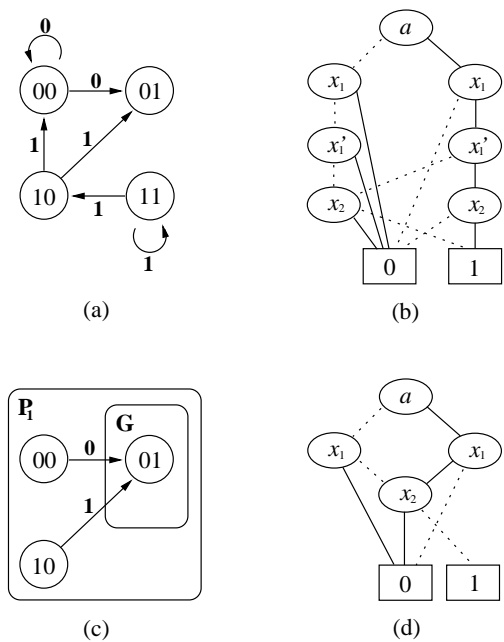

Figure 4: A planning domain represented as an NFA is shown in (a). States are defined by boolean state variables $x_1$ and $x_2$, and the action input to the NFA is given by the boolean variable $a$. The symbolic representation of the transition relation of the NFA is shown in (b). In (c), $P_1$ is the set of state action pairs for which, execution of the action can lead to the goal. The symbolic representation of $P_1$ is shown in (d). It is obtained from the transition relation by restricting the next state to 01.

Assume that the state 01 is a goal state $G$. A key operation, when generating a universal plan for achieving $G$, is to find all the state action pairs $(s, a)$ such that $G$ can be reached from $s$ by executing $a$. This set is labeled $P_1$ in Figure 4c. To find $P_1$ from $T$ we constrain $x_1'$ to *False* and $x_2'$ to *True* in $T$. This reduces $T$ to the OBDD shown in Figure 4d. The resulting OBDD represents $P_1$ with the states described in the current state variables $x_1$ and $x_2$. Logically we performed the operation $\exists x_1', x_2'. \neg x_1' \wedge x_2' \wedge T$ to obtain the OBDD representing $P_1$.

---

2. Another notation like $x_t$ and $x_{t+1}$ could have been used for current and next state variables. We have chosen the quote notation because it is the common notation in model checking.

## 4. The *NADL* Description Language

In this section, we first discuss the properties of *NADL* based on an informal definition of the language and a domain encoding example. We then describe the formal syntax and semantics of *NADL*.

An *NADL* domain description consists of: a definition of *state variables*, a description of *system* and *environment agents*, and a specification of an *initial* and *goal condition*.

The set of state variable assignments defines the state space of the domain. An agent's description is a set of *actions*. The agents change the state of the world by performing actions that are assumed to be executed synchronously and to have a fixed and equal duration. At each step, all of the agents perform exactly one action, and the resulting action tuple is a *joint action*. The system agents model the behavior of the agents controllable by the planner, while the environment agents model the uncontrollable world. A valid domain description requires that the system and environment agents constrain a disjoint set of variables.

An action has three parts: a set of *state variables*, a *precondition* formula, and an *effect* formula. Intuitively the action takes responsibility of constraining the values of the state variables in the next state. It further has exclusive access to these variables during execution. In order for the action to be applicable, the precondition formula must be satisfied in the current state. The effect of the action is defined by the effect formula that must be satisfied in the next state. To allow conditional effects, the effect expression can refer to both current and next state variables, where the next state variables need to be a part of the set of constrained variables of the action. All next state variables not constrained by any action in a joint action maintain their value.

The initial and goal condition are formulas that must be satisfied in the initial state and the final state, respectively.

There are two causes for non-determinism in *NADL* domains: (1) actions not restricting all their constrained variables to a specific value in the next state, and (2) the non-deterministic selection of environment actions.

A simple example of an *NADL* domain description is shown in Figure 5. The domain describes a planning problem for Schoppers' (1987) robot-baby domain. The domain has two state variables: a numerical one, position *pos* with range $\{0, 1, 2, 3\}$ and a propositional one, *robot_works*. The robot is the only system agent and it has two actions *Lift-Block* and *Lower-Block*. The *Lift-Block* (and *Lower-Block*) action has a conditional effect described by an if-then-else operator ($\rightarrow$): if *robot_works* is true then *Lift-Block* increases the block position by one else the block position is unchanged. The baby is the only environment agent and it has one action *Hit-Robot*. Because each agent must perform exactly one action at each step, there are two joint actions (*Lift-Block,Hit-Robot*) and (*Lower-Block,Hit-Robot*). Initially *robot_works* is assumed to be true, the robot is assumed to hold a block at Position 0, and its task is to lift it up to Position 3.

The variable *robot_works* can be made false by the baby. The baby's action *Hit-Robot* is non-deterministic, as it only constrains *robot_works* by the effect expression $\neg robot\_works \Rightarrow$

**variables**
  **nat**(4) *pos*
  **bool** *robot_works*
**system**
  **agt:** Robot
    Lift-Block
      **con:** *pos*
      **pre:** *pos* < 3
      **eff:** *robot_works* → *pos'* = *pos* + 1, *pos'* = *pos*
    Lower-Block
      **con:** *pos*
      **pre:** *pos* > 0
      **eff:** *robot_works* → *pos'* = *pos* − 1, *pos'* = *pos*
**environment**
  **agt:** Baby
    Hit-Robot
      **con:** *robot_works*
      **pre:** *true*
      **eff:** ¬*robot_works* ⇒ ¬*robot_works'*
**initially**
  *pos* = 0 ∧ *robot_works*
**goal**
  *pos* = 3

Figure 5: An *NADL* domain description.

¬*robot_works'*. Thus, when *robot_works* is true in the current state, the effect expression of *Hit-Robot* does not apply, and *robot_works* can either be true or false in the next state. On the other hand, if *robot_works* is false in the current state, *Hit-Robot* keeps it false in the next state. The *Hit-Robot* action models an aspect of the environment not controlled by the robot agent, in this case a baby, by its effects on *robot_works*. In the example above, *robot_works* stays false when it has become false, reflecting that the robot cannot spontaneously be fixed by a hit of the baby, or any other action in the environment.

An NFA representing the domain is shown in Figure 6. The calculation of the next state value of *pos* in the *Lift-Block* action shows that numerical variables can be updated by an arithmetic expression on the current state variables. The update expression of *pos* and the use of the if-then-else operator further demonstrate the advantage of using explicit references to current state and next state variables in effect expressions. *NADL* does not restrict the representation by enforcing a structure separating current state and next state expressions. The if-then-else operator has been added to support complex conditional effects that often are efficiently and naturally represented as a set of nested if-then-else operators.

The explicit representation of constrained state variables enables any non-deterministic or deterministic effect of an action to be represented, as the constrained variables can be assigned to any value in the next state that satisfies the effect formula. It further turns out to have a clear intuitive meaning, as the action takes the "responsibility" of specifying the values of the constrained variables in the next state.

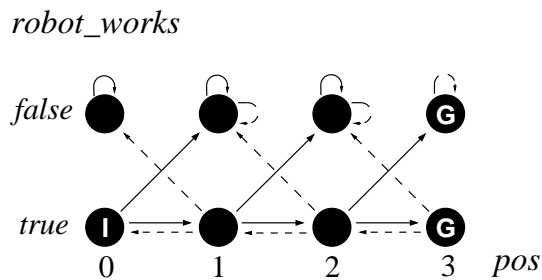

Figure 6: The NFA of the robot-baby domain (see Figure 5). There is one propositional and one numerical state variable: *robot_works* and *pos*. The (*Lift-Block,Hit-Robot*) and (*Lower-Block,Hit-Robot*) joint actions are drawn with solid and dashed arrows, respectively. States marked with "I" and "G" are initial and goal states.

Compared to the action description language $\mathcal{A}$ (Gelfond & Liftschitz, 1993) and $\mathcal{AR}$ that are the only prior languages used for OBDD-based planning (Di Manzo et al., 1998; Cimatti et al., 1998a, 1998b, 1997), *NADL* introduces an explicit environment model, a multi-agent decomposition, and numerical state variables. It can further be shown that *NADL* can be used to model any domain that can be modelled by $\mathcal{AR}$ (see Appendix A).

The concurrent actions in *NADL* are assumed to be synchronously executed and to have fixed and equal duration. A general representation allowing partially overlapping actions and actions with different durations has been avoided, as it requires more complex temporal planning (see e.g., O-PLAN or PARCPLAN, Currie & Tate, 1991; Lever & Richards, 1994). Our joint action representation has more resemblance with $\mathcal{A_C}$ (Baral & Gelfond, 1997) and $\mathcal{C}$ (Giunchiglia & Lifschitz, 1998), where sets of actions are performed at each time step. In contrast to these approaches, though, we model multi-agent domains.

An important issue to address when introducing concurrent actions is synergetic effects between simultaneously executing actions (Lingard & Richards, 1998). A common example of destructive synergetic effects is when two or more actions require exclusive use of a single resource or when two actions have inconsistent effects like $pos' = 3$ and $pos' = 2$.

In *NADL* actions cannot be performed concurrently in the following two conditions: 1) they have inconsistent effects, or 2) they constrain an overlapping set of state variables. The first condition is due to the fact that state knowledge is expressed in a monotonic logic that cannot represent inconsistent knowledge. The second condition addresses the problem of sharing resources. Consider for example two agents trying to drink the same glass of water. If only the first condition defined interfering actions, both agents could simultaneously empty the glass, as the effect *glass_empty* of the two actions would be consistent. With the second condition added, these actions are interfering and cannot be performed concurrently.

The current version of *NADL* only avoids destructive synergetic effects. It does not include ways of representing constructive synergetic effects between simultaneous acting agents. A constructive synergetic effect is illustrated in Baral and Gelfond (1997), where an agent spills soup from a bowl when trying to lift it up with one hand, but not when lifting it up with both hands. In $\mathcal{C}$ and $\mathcal{A_C}$ this kind of synergetic effects can be represented by explicitly stating the effect of a compound action. A similar approach could be used in *NADL* but is currently not supported.

## 4.1 Syntax

Formally, an *NADL* description is a 7-tuple $D = (SV, S, E, Act, d, I, G)$, where:

- $SV = PVar \cup NVar$ is a finite set of state variables comprised of a finite set of propositional variables, *PVar*, and a finite set of numerical variables, *NVar*.

- $S$ is a finite, nonempty set of system agents.

- $E$ is a finite set of environment agents.

- *Act* is a set of action descriptions $(c, p, e)$ where $c$ is the set of state variables constrained by the action, $p$ is a precondition state formula in the set *SForm* and $e$ is an effect formula in the set *Form*. Thus $(c, p, e) \in Act \subset 2^{SV} \times SForm \times Form$. The sets *SForm* and *Form* are defined below.

- $d : Agt \rightarrow 2^{Act}$ is a function mapping agents $(Agt = S \cup E)$ to their actions. Because an action belongs to exactly one agent, $d$ must satisfy the following conditions:

$$\bigcup_{\alpha \in Agt} d(\alpha) = Act$$

$$\forall \alpha_1, \alpha_2 \in Agt . \alpha_1 \neq \alpha_2 \Rightarrow d(\alpha_1) \cap d(\alpha_2) = \emptyset$$

- $I \in SForm$ is the initial condition.

- $G \in SForm$ is the goal condition.

For a valid domain description, we require that actions of system agents are independent of actions of environment agents:

$$\bigcup_{\substack{e \in E \\ a \in d(e)}} c(a) \ \cap \ \bigcup_{\substack{s \in S \\ a \in d(s)}} c(a) = \emptyset,$$

where $c(a)$ is the set of constrained variables of action $a$.

The set of formulas *Form* is constructed from the following alphabet of symbols:

- A finite set of current state $v$ and next state $v'$ variables, where $v, v' \in SV$.

- The natural numbers $\mathbf{N}$.

- The arithmetic operators $+, -, /, *$ and *mod*.

- The relation operators $>, <, \leq, \geq, =$ and $\neq$.

- The boolean operators $\neg, \vee, \wedge, \Rightarrow, \Leftrightarrow$ and $\rightarrow$.

- The special symbols *true*, *false*, parentheses and comma.

The set of arithmetic expressions is constructed from the following rules:

1. Every numerical state variable $v \in NVar$ is an arithmetic expression.

2. A natural number is an arithmetic expression.

3. If $e_1$ and $e_2$ are arithmetic expressions and $\oplus$ is an arithmetic operator, then $e_1 \oplus e_2$ is an arithmetic expression.

Finally, the set of formulas *Form* is generated by the rules:

1. *true* and *false* are formulas.

2. Propositional state variables $v \in PVar$ are formulas.

3. If $e_1$ and $e_2$ are arithmetic expressions and $\mathcal{R}$ is a relation operator, then $e_1 \mathcal{R} e_2$ is a formula.

4. If $f_1$, $f_2$ and $f_3$ are formulas, so are $(\neg f_1)$, $(f_1 \vee f_2)$, $(f_1 \wedge f_2)$, $(f_1 \Rightarrow f_2)$, $(f_1 \Leftrightarrow f_2)$ and $(f_1 \to f_2, f_3)$.

Parentheses have their usual meaning and operators have their usual priority and associativity with the if-then-else operator "$\to$" given lowest priority.

$SForm \subset Form$ is a subset of the formulas only referring to current state variables. These formulas are called *state formulas*.

## 4.2 Semantics

All of the symbols in the alphabet of formulas have their usual meaning with the if-then-else operator $f_1 \to f_2, f_3$ being an abbreviation for $(f_1 \wedge f_2) \vee (\neg f_1 \wedge f_3)$. Each numerical state variable $v \in NVar$ has a finite range $rng(v) = \{0, 1, \cdots, t_v\}$, where $t_v > 0$.

The formal semantics of a domain description $D = (SV, S, E, Act, d, I, G)$ is given in terms of an NFA $M$:

**Definition 1 (NFA)** *A Non-deterministic Finite Automaton is a 3-tuple, $M = (Q, \Sigma, \delta)$, where $Q$ is a set of states, $\Sigma$ is a set of input values, and $\delta : Q \times \Sigma \to 2^Q$ is a next state function.*

In the following construction of $M$, we express the next state function as a transition relation. Let $\mathcal{B}$ denote the set of boolean values $\{True, False\}$. Further, let the *characteristic function* $A \colon B \to \mathcal{B}$ associated to a set $A \subseteq B$ be defined by $A(x) = (x \in A)$.[3] Given an NFA $M$ we define its *transition relation* $T \subseteq Q \times \Sigma \times Q$ as a set of triples with characteristic function $T(s, i, s') = (s' \in \delta(s, i))$.

The set of states $Q$ of $M$ equals the set of all possible variable assignments $Q = (PVar \to \mathcal{B}) \times (Nvar \to \mathbf{N})$. The input $\Sigma$ of $M$ is the set of joint actions of system agents represented as sets. That is, $\{a_1, a_2, \cdots, a_{|S|}\} \in \Sigma$ if and only if $(a_1, a_2, \cdots, a_{|S|}) \in \prod_{\alpha \in S} d(\alpha)$, where $|S|$ denotes the number of elements in $S$.

---

3. Note: the characteristic function has the same name as the set.

We define the transition relation $T : Q \times \Sigma \times Q \to \mathcal{B}$ of $M$ by:

$$T(s, i, s') = \exists j \in J \,.\, i \subseteq j \wedge t(s, j, s'),$$

where $t : Q \times J \times Q \to \mathcal{B}$ is the transition relation for joint actions $J$ of both system and environment agents. The existential quantification makes the actions of environment agents uncontrollable, since $T(s, i, s')$ is true, if there exists some joint action of environment agents $i_e$ such that the combined joint action $j = i \cup i_e$ makes $t(s, j, s')$ true.

The transition relation $t$ is a conjunction of three relations $A$, $F$ and $I$, $t(s, j, s') = A(s, j, s') \wedge F(s, j, s') \wedge I(j)$. Given an action $a = (c, p, e)$, a current state $s$ and next state $s'$, let $P_a(s)$ and $E_a(s, s')$ denote the value of the precondition formula $p$ and effect formula $e$ of $a$, respectively.

$A : Q \times J \times Q \to \mathcal{B}$ is then defined by:

$$A(s, j, s') = \bigwedge_{a \in j} \Big( P_a(s) \wedge E_a(s, s') \Big).$$

$A$ defines the constraints on the current state and next state of joint actions. $A$ further ensures that actions with inconsistent effects cannot be performed concurrently, as $A$ reduces to false if any pair of actions in a joint action has inconsistent effects. Thus, $A$ also states the first condition (see Section 4) for avoiding interference between concurrent actions.

$F : Q \times J \times Q \to \mathcal{B}$ is a frame relation ensuring that unconstrained variables maintain their value. Let $c(a)$ denote the set of constrained variables of action $a$. We then have:

$$F(s, j, s') = \bigwedge_{v \in SV \setminus C} (v = v'),$$

where $C = \bigcup_{a \in j} c(a)$.

$I : J \to \mathcal{B}$ ensures that concurrent actions constrain a non overlapping set of variables and thus states the second condition for avoiding interference between concurrent actions:

$$I(j) = \bigwedge_{(a_1, a_2) \in j^2} \Big( c(a_1) \cap c(a_2) = \emptyset \Big),$$

where $j^2$ denotes the set $\{(a_1, a_2) \,|\, (a_1, a_2) \in j \times j \wedge a_1 \neq a_2\}$.

## 5. OBDD Representation of NADL Descriptions

To build an OBDD $\tilde{T}$ representing the transition relation $T(s, i, s')$ of the NFA of a domain description $D = (SV, S, E, Act, d, I, G)$, we must define a set of boolean variables to represent the current state $s$, the joint action input $i$, and the next state $s'$. As in Section 4.2 we first build a transition relation with the joint actions of both system and environment agents as input and then reduce it to a transition relation with only joint actions of system agents as input.

Joint action inputs are represented in the following way: assume action $a$ is identified by a number $p$ and can be performed by agent $\alpha$. $a$ is then defined to be the action of agent $\alpha$, if the number expressed in binary by a set of boolean variables $A_\alpha$, used to represent the actions of $\alpha$, is equal to $p$. Propositional state variables are represented by a single boolean variable, while numerical state variables are represented in binary by a set of boolean variables.

Let $A_{e_1}, \ldots, A_{e_{|E|}}$ and $A_{s_1}, \ldots, A_{s_{|S|}}$ denote sets of boolean variables used to represent the joint action of environment and system agents. Further, let $x_{v_j}^k$ and $x'^k_{v_j}$ denote the $k$th boolean variable used to represent state variable $v_j \in SV$ in the current and next state. The boolean variables are ordered with the input variables first, followed by an interleaving of the boolean variables of current state and next state variables:

$$A_{e_1} \prec \cdots \prec A_{e_{|E|}} \prec A_{s_1} \prec \cdots \prec A_{s_{|S|}}$$
$$\prec x_{v_1}^1 \prec x'^1_{v_1} \prec \cdots \prec x_{v_1}^{m_1} \prec x'^{m_1}_{v_1}$$
$$\cdots$$
$$\prec x_{v_n}^1 \prec x'^1_{v_n} \prec \cdots \prec x_{v_n}^{m_n} \prec x'^{m_n}_{v_n},$$

where $m_i$ is the number of boolean variables used to represent state variable $v_i$ and $n$ is equal to $|SV|$. The construction of $\tilde{T}$ is quite similar to the construction of $T$ in Section 4.2. An OBDD representing a logical expression is built in the standard way (Bryant, 1986). Arithmetic expressions are represented as lists of OBDDs defining the corresponding binary number. They collapse to single OBDDs when related by arithmetic relations.

To build an OBDD $\tilde{A}$ defining the constraints of the joint actions, we need to refer to the values of the boolean variables representing the actions. Let $i(\alpha)$ be the function that maps an agent $\alpha$ to the value of the boolean variables representing its action and let $b(a)$ be the identifier value of action $a$. Further let $\tilde{P}(a)$ and $\tilde{E}(a)$ denote OBDD representations of the precondition and effect formula of an action $a$. $\tilde{A}$ is then given by:

$$\tilde{A} = \bigwedge_{\substack{\alpha \in Agt \\ a \in d(\alpha)}} \Big( i(\alpha) = b(a) \Rightarrow \tilde{P}(a) \wedge \tilde{E}(a) \Big).$$

Note that logical operators now denote the corresponding OBDD operators.

An OBDD representing the frame relation $\tilde{F}$ changes in a similar way:

$$\tilde{F} = \bigwedge_{v \in SV} \Big( ( \bigwedge_{\substack{\alpha \in Agt \\ a \in d(\alpha)}} (i(\alpha) = b(a) \Rightarrow v \notin c(a))) \Rightarrow s'_v = s_v \Big),$$

where $c(a)$ is the set of constrained variables of action $a$ and $s_v = s'_v$ expresses that all current and next state boolean variables representing $v$ are pairwise equal. The expression $v \notin c(a)$ evaluates to $True$ or $False$ and is represented by the OBDD for $True$ or $False$. The action interference constraint $\tilde{I}$ is given by:

$$\tilde{I} = \bigwedge_{\substack{(\alpha_1, \alpha_2) \in S^2 \\ (a_1, a_2) \in c(\alpha_1, \alpha_2)}} \Big( i(\alpha_1) = b(a_1) \Rightarrow i(\alpha_2) \neq b(a_2) \Big) \wedge$$

$$\bigwedge_{\substack{(\alpha_1, \alpha_2) \,\in\, E^2 \\ (a_1, a_2) \,\in\, c(\alpha_1, \alpha_2)}} \Big( i(\alpha_1) = b(a_1) \Rightarrow i(\alpha_2) \neq b(a_2) \Big),$$

where $c(\alpha_1, \alpha_2) = \{(a_1, a_2) \,|\, (a_1, a_2) \in d(\alpha_1) \times d(\alpha_2) \wedge c(a_1) \cap c(a_2) \neq \emptyset\}$.

Finally the OBDD representing the transition relation $\tilde{T}$ is the conjunction of $\tilde{A}$, $\tilde{F}$ and $\tilde{I}$ with action variables of the environment agents existentially quantified:

$$\tilde{T} = \exists A_{e_1}, \cdots, A_{e_{|E|}} \,.\, \tilde{A} \wedge \tilde{F} \wedge \tilde{I}.$$

## 5.1 Partitioning the Transition Relation

The algorithms we use for generating universal plans all consist of a backward search from the states satisfying the goal condition to the states satisfying the initial condition. Empirical studies in model checking have shown that the most complex operation for this kind of algorithms normally is to find the *preimage* of a set of visited states $V$ (Ranjan et al., 1995).

**Definition 2 (Preimage)** *Given an NFA $M = (Q, \Sigma, \delta)$ and a set of states $V \subseteq Q$, the preimage of $V$ is the set of states $\{s \,|\, s \in Q \wedge \exists i \in \Sigma, s' \in \delta(s, i) \,.\, s' \in V\}$.*

A preimage is said to exist, if it is nonempty. Note that states already belonging to $V$ can also be a part of the preimage of $V$. Assume that the set of visited states are represented by an OBDD expression $\tilde{V}$ on next state variables and that, for iteration purposes, we want to generate the preimage $\tilde{P}$ also expressed in next state variables. For a monolithic transition relation $\tilde{T}$ we then calculate:

$$\begin{aligned} \tilde{U} &= (\exists \vec{x}' \,.\, \tilde{T} \wedge \tilde{V})[\vec{x}/\vec{x}'] \\ \tilde{P} &= \exists \vec{i} \,.\, \tilde{U} \end{aligned}$$

where $\vec{i}$, $\vec{x}$ and $\vec{x}'$ denote input, current state and next state variables, and $[\vec{x}/\vec{x}']$ denotes the substitution of current state variables with next state variables. The set expressed by $\tilde{U}$ consists of state input pairs $(s, i)$, for which the state $s$ belongs to the preimage of $V$ and the input $i$ may cause a transition from $s$ to a state in $V$. The input of an NFA representing a planning domain is a set of actions. Thus, for a planning domain the elements in $\tilde{U}$ are state-action pairs. The generated universal plans of the universal planning algorithms presented in the next section are sets of these state-action pairs. We refer to the state-action pairs as *state-action rules*, because they associate states to actions that can be performed in these states.

The OBDD representing the transition relation $\tilde{T}$ and the set of visited states $\tilde{V}$ tends to be large, and a more efficient computation can be obtained by performing the existential quantification of next state variables early in the calculation (Burch et al., 1991; Ranjan et al., 1995). To do this, the transition relation has to be split into a conjunction of partitions $\tilde{T} = \tilde{T}_1 \wedge \cdots \wedge \tilde{T}_n$ allowing the modified calculation:

$$\begin{aligned} \tilde{U} &= (\exists \vec{x}'_n \,.\, \tilde{T}_n \wedge \cdots (\exists \vec{x}'_2 \,.\, \tilde{T}_2 \wedge (\exists \vec{x}'_1 \,.\, \tilde{T}_1 \wedge \tilde{V})) \cdots)[\vec{x}/\vec{x}'] \\ \tilde{P} &= \exists \vec{i} \,.\, \tilde{U} \end{aligned}$$

That is, $\bar{T}_1$ can refer to all variables, $\bar{T}_2$ can refer to all variables except $\bar{x}'_1$, $\bar{T}_3$ can refer to all variables except $\bar{x}'_1$ and $\bar{x}'_2$ and so on.

As shown by Ranjan et al. (1995) the computation time used to calculate the preimage is a convex function of the number of partitions. The reason for this is that, for some number of partitions, a further subdivision of the partitions will not reduce the total complexity, because the complexity introduced by the larger number of OBDD operations is higher than the reduction of the complexity of each OBDD operation.

The representation of the logical expression for each relation $A$, $F$ and $I$ has been carefully chosen such that it consists of a conjunction of subexpressions that only refer to a small subset of next state variables. This representation allows us to sort out the subexpressions in conjunctive partitions with near optimal sizes that satisfy the above requirements.

## 6. OBDD-based Universal Planning Algorithms

We first describe two prior algorithms for OBDD-based universal planning and discuss which kind of domains they are suitable for. We then present a new algorithm called *optimistic planning* that is suitable for some domains not covered by the prior algorithms.

The three universal planning algorithms discussed are all based on an iteration of preimage calculations. The iteration corresponds to a parallel backward breadth-first search starting at the goal states and ending when all initial states are included in the set of visited states (see Figure 7). The main difference between the algorithms is the way the preimage is defined.

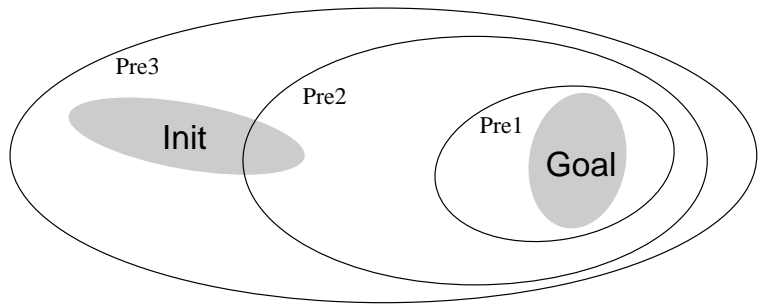

Figure 7: An illustration of the parallel backward breadth-first search used by OBDD-based universal planning algorithms, computing preimages Pre1, Pre2 and Pre3.

### 6.1 Strong and Weak Preimages

Cimatti et al. (1998a) introduces two different kinds of preimages called *strong* and *weak preimages*. A strong preimage is defined by:

**Definition 3 (Strong Preimage)** *Given an NFA $M = (Q, \Sigma, \delta)$ and a set of states $V \subseteq Q$, the strong preimage of $V$ is the set of states $\{s \mid s \in Q \land \exists i \in \Sigma . \delta(s, i) \subseteq V\}$.*

Thus, for a state $s$ belonging to the strong preimage of a set of states $V$, there exists at least one action $i$ where all the transitions from $s$ associated with $i$ lead into $V$. Consider the example shown in Figure 8. The dots and arrows in this figure denote states and transitions for an NFA with a single non-deterministic action. For the set of states GS shown in the figure, the three states having a transition into GS are the strong preimage of GS (indicated by a solid ellipse and labelled pre1), as all transitions from these states lead into GS.

A **weak preimage** is equal to an ordinary preimage as defined in Definition 2. Thus, in Figure 8 all the strong preimages are also weak preimages, but the preimages shown by dashed ellipses are only weak preimages, as the dashed transitions do not satisfy the strong preimage definition.

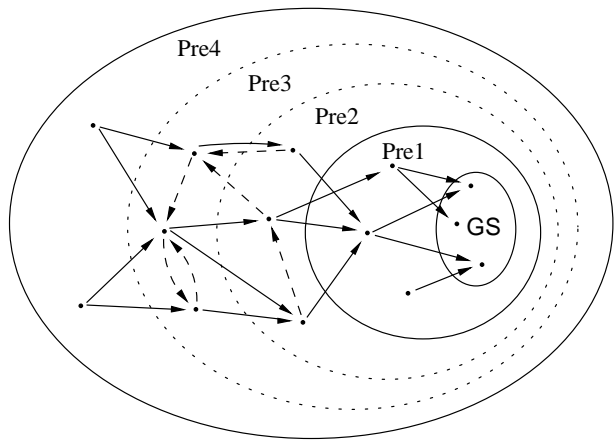

Figure 8: Strong and weak preimage calculations. Solid ellipses denote preimages that are both strong and weak, while dashed ellipses denote preimages that are only weak. Only one action is assumed to exist in the domain. Transitions causing a state to belong to a weak preimage rather than a strong preimage are drawn dashed. The set of goal states is marked "GS".

## 6.2 Strong and Strong Cyclic Planning

A strong or strong cyclic plan is the union of the state-action rules $U$ found when calculating the preimages necessary for covering the set of initial states ($U$ is defined in Section 5).

Strong planning only considers strong preimages. If a sequence of strong preimages starting at the set of goal states can be calculated, such that the set of initial states is covered, strong planning succeeds and returns the universal plan consisting of the union of all the state-action rules of the calculated strong preimages. Otherwise it fails (Cimatti et al., 1998b).

Consider the example in Figure 8. As depicted in the figure, a strong preimage can be found in the first preimage calculation, but only a weak preimage can be found in the second calculation. Thus, strong planning only succeeds in this example, if the set of initial states is covered by the first preimage and the set of goal states $GS$.

Strong planning is complete with respect to strong solutions. If a strong plan exists for some planning problem the strong planning algorithm will return it, otherwise, it returns that no solution exists. Strong planning is also optimal due to the breadth-first search. Thus, a strong plan with the fewest number of steps in the worst case is returned.

Strong cyclic planning is a relaxed version of strong planning, because it also considers weak preimages. Strong cyclic planning finds a strong plan, if it exists. Otherwise, when unable to find a strong preimage the algorithm adds a weak preimage. It then tries to prune this preimage by removing all states that have transitions leading out of the preimage and the set of visited states $V$. If it succeeds, the remaining states in the preimage are added to $V$ and it again tries to add strong preimages. If it fails, it adds a new, weak preimage and repeats the pruning process (Cimatti et al., 1998a).

Consider again the example in Figure 8. The shown sequence of preimage calculations could have been computed by the strong cyclic planning algorithm. The algorithm prefers strong preimages, if they exist, so the first added preimage (Pre1) is strong. No strong second preimage exists and the weak preimage (Pre2) cannot be pruned to only contain states not having transitions leading out of the preimage and the set of visited states. Thus, the strong cyclic algorithm looks for another weak preimage. This preimage (Pre3) has no outgoing transitions, which means that the sequence of weak preimages can be terminated and the algorithm can return to look for strong preimages (Pre4). If the set of initial states after adding preimage Pre4 covers the set of initial states the algorithm succeeds, otherwise it continues until either no strong or pruned weak preimage can be found (in which case the algorithm fails) or the set of visited states covers the set of initial states (in which case the algorithm succeeds).

A strong cyclic plan only guarantees progress towards the goal in the strong parts. In the weak parts, cycles can occur. To ensure that the plan length is finite, we must assume that transitions leading out of the weak parts eventually will be taken. The algorithm is complete with respect to strong solutions, as a strong solution will be returned, if it exists.

## 6.3 Strengths and Limitations of Strong and Strong Cyclic Planning

An important reason for studying universal planning is that universal planning algorithms can provide state-action rules to completely handle a non-deterministic environment. Thus, if a plan exists for painting the floor, an agent executing a universal plan will always avoid painting itself into the corner or reach any other unrecoverable dead-end. Strong planning and strong cyclic planning algorithms contribute by providing complete OBDD-based algorithms for universal planning.

Unfortunately, real-world domains can have dead-ends that are not always avoidable. Consider, for example, Schoppers' robot-baby domain described in Section 4. As depicted in Figure 6, no universal plan represented by a set of state-action rules can guarantee the goal to be reached in a finite or infinite number of steps, as all relevant actions may lead to an unrecoverable dead-end.

A more interesting example is how to generate a universal plan for a system that can be in a bad state, good state or an unrecoverable failed state (dead-end). Assume that actions can be executed that can bring the system from any bad state to a good state, but environment actions unfortunately can also make the system stay in a bad state or even change to an unrecoverable failed state (see Figure 9). No strong nor strong cyclic solution can be found, because an unrecoverable state can be reached from any initial state. An example of such a domain (a power plant) is studied in Section 7.1.2.

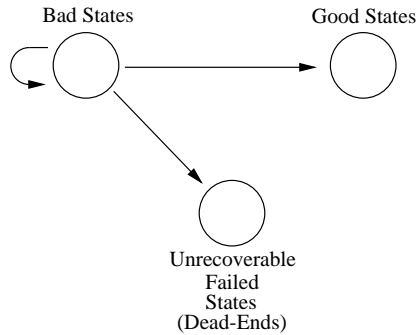

Figure 9: Abstract description of the NFA of a system with unrecoverable states.

Another limitation of strong and strong cyclic planning is the inherent pessimism of these algorithms. Consider for example the domain (Domain 1) illustrated in Figure 10. The domain consists of $n + 1$ states and two different actions (dashed and solid).

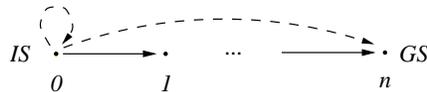

Figure 10: Domain 1. The NFA of a domain with two actions (drawn as solid and dashed arrows) illustrating the possible loss of short plan lengths when preferring strong solutions. IS and GS are the initial and goal state, respectively.

The strong cyclic algorithm returns a strong plan $\{(0, solid), (1, solid), \cdots, (n-1, solid)\}$. This plan would have a best and worst-case length of $n$. But a strong cyclic plan $\{(0, dashed), (n-1, solid)\}$ also exists and could be preferable because the best-case length of 1 of the cyclic solution may have a much higher probability than the infinite worst-case length. Strong cyclic planning will always prefer to return a strong plan, if it exists, even though a strong cyclic plan may exist with a shorter, best-case plan length.

By adding an unrecoverable dead-end for the dashed action and making solid actions non-deterministic (see Domain 2, Figure 11), strong cyclic planning now returns the strong cyclic plan $\{(0, solid), (1, solid), \cdots, (n-1, solid)\}$. But we might still be interested in the plan $\{(0, dashed), (n-1, solid)\}$ even though the goal is not guaranteed to be achieved.

## 6.4 Optimistic Planning

The analysis in the previous section shows that there are domains and planning problems for which we may want to use a fully relaxed algorithm that always includes the best-case plan and returns a solution even if it includes dead-ends that cannot be guaranteed to be avoided. We introduce an algorithm similar to the strong planning algorithm that adds an ordinary preimage in each iteration has these properties. Because state-action rules leading to unrecoverable dead-ends may be added to the universal plan, we call this algorithm *optimistic*

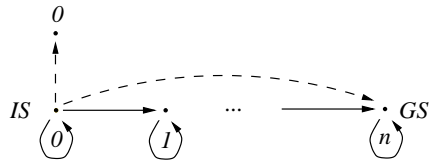

Figure 11: Domain 2. The NFA of a domain with two actions (drawn as solid and dashed arrows) illustrating the possible loss short plan lengths when preferring strong cyclic solutions. IS and GS are the initial and goal state, respectively.

*planning*. The algorithm is shown in Figure 12. The function Preimage(*VisitedStates*) returns the set of state-action rules $U$ associated with the preimage of the visited states. Prune(*StateActions*, *VisitedStates*) removes the state-action rules, where the state already is included in the set of visited states, and StatesOf(*PrunedStateActions*) returns the set of states of the pruned state-action rules. UMOP includes the optimistic planning algorithm. The optimistic planning algorithm is incomplete with respect to strong solutions, because it

> **procedure** OptimisticPlanning(*Init*, *Goal*)
>     *VisitedStates* := *Goal*
>     *UniversalPlan* := ∅
>     **while** (*Init* ⊄ *VisitedStates*)
>         *StateActions* := Preimage(*VisitedStates*)
>         *PrunedStateActions* := Prune(*StateActions*, *VisitedStates*)
>         **if** *PrunedStateActions* ≠ ∅ **then**
>             *UniversalPlan* := *UniversalPlan* ∪ *PrunedStateActions*
>             *VisitedStates* := *VisitedStates* ∪ StatesOf(*PrunedStateActions*)
>         **else**
>             **return** "No optimistic plan exists"
>     **return** *UniversalPlan*

Figure 12: The optimistic planning algorithm.

does not necessarily return a strong solution, if one exists. Intuitively, optimistic planning only guarantees that there exists some effect of a plan action leading to the goal, where strong planning guarantees that all effects of plan actions lead to the goal.

The purpose of optimistic planning is not to substitute strong or strong cyclic planning. These algorithms should be used in domains where strong or strong cyclic plans can be found and goal achievement has the highest priority. Optimistic planning might be the better choice in domains where goal achievement cannot be guaranteed or the shortest plan should be included in the universal plan.

Consider again, as an example, the robot-baby domain described in Section 4. For this problem the optimistic solution makes the robot try to lift the block when the position of the block is less than 3 and the robot is working. This seems to be the only reasonable strategy, even though no guarantee for goal achievement can be given. It is worthwhile

| Domain | Strong | | Strong Cyclic | | Optimistic | |
|---|---|---|---|---|---|---|
| | best | worst | best | worst | best | worst |
| 1 | $n$ | $n$ | 1 | $\infty$ | 1 | $\infty$ |
| 2 | - | - | $n$ | $\infty$ | 1 | $\infty_D$ |

Table 1: The best and worst-case plan length of possible strong, strong cyclic and optimistic plans in Domains 1 and 2 (see Figures 10 and 11).(-) means that no solution exists. ($\infty_D$) indicates that the plan length is infinite, and an unrecoverable dead-end is reached.

constructing an optimistic plan for this kind of domains since the alternative is no plan at all.

A similar optimistic plan is generated for the domain shown in Figure 9. For all bad states, the optimistic plan associates an action that brings the system to a good state in one step. This continues as long as the environment keeps the system in a bad state. Because no strategy can be used to prevent the environment from bringing the system to an unrecoverable dead-end, the optimistic solution is quite sensible.

For Domains 1 and 2 shown in Figures 10 and 11, optimistic planning returns a universal plan $\{(0, dotted), (n-1, solid)\}$. For both domains this is a universal plan with the shortest best-case length. Compared to the strong cyclic solution the cost in the first domain is that the plan may have an infinite length, while the cost in the second domain is that a dead-end may be reached. The results of strong, strong cyclic, and optimistic planning in Domains 1 and 2 are summarized in Table 1.

## 7. Empirical Results

The input to UMOP is an *NADL* description[4] and a specification of which planning algorithm to use. This description is then converted to a set of OBDDs representing the partitioned transition relation as described in Section 5. The OBDD representation is used by a set of planning algorithms to generate a plan. The output of UMOP is a universal plan or sequential plan depending on the planning algorithm. A universal plan is represented by an OBDD. It defines for each domain state a set of joint actions that the system agents must execute synchronously in order to achieve the goal. The implemented planning algorithms are:

1. Strong planning.

2. Strong cyclic planning.

3. Optimistic planning.

4. Classical deterministic planning.

---

4. The *NADL* description accepted by the current implementation includes all logical operators but only the arithmetic operators $+$ and $-$. An implementation of the remaining operators is straightforward and is part of our current work.

Deterministic planning can be viewed as a special case of non-deterministic planning. In UMOP, we used the optimistic planning algorithm for the backward search of classical deterministic planning. (The strong or strong cyclic algorithm could also have been used, as all the described non-deterministic algorithms behave similarly in deterministic domains.) The only new feature of the deterministic algorithm is that a sequential plan is generated from the universal plan by choosing an initial state and iteratively adding an action from the universal plan until a goal state is reached. The deterministic planning algorithm has been implemented to verify the performance of UMOP compared to other classical planners. It has not been our intention in this work, though, to develop a fast OBDD-based classical planning algorithm like Di Manzo et al. (1998). Our main interest is non-deterministic, multi-agent universal planning.

The UMOP planning system is implemented in C/C++ and uses the BUDDY package (Lind-Nielsen, 1999) for OBDD manipulations. During planning the dynamic variable reordering facility of the BUDDY package is used to find a better ordering of the OBDD variables.

In the following four subsections we present results obtained with the UMOP planning system in nine different domains ranging from deterministic and single-agent with no environment actions to non-deterministic and multi-agent with complex environment actions. All experiments were carried out on a 450 MHz Pentium PC with 1 GByte RAM running Red Hat Linux 4.2. A more detailed description of the experiments including the complete description of the *NADL* domains can be found in Jensen (1999).

## 7.1 Non-Deterministic Domains

We first test UMOP's performance for some of the non-deterministic domains solved by MBP. Next, we present a power plant domain and finally, we show results from a multi-agent soccer domain.

### 7.1.1 Domains Tested by Mbp

One of the domains solved by MBP is a non-deterministic transportation domain. The domain consists of a set of locations and a set of actions like drive-truck, drive-train and fly to move between the locations. Non-determinism is caused by non-deterministic actions (e.g., after a drive action a truck may or may not have fuel left) and environmental changes (e.g., fog at airports, Cimatti et al., 1998a). We defined the two domain examples tested by MBP for strong and strong cyclic planning in *NADL* and ran UMOP using strong and strong cyclic planning. Both examples were solved in less than 0.05 seconds. Similar results were obtained with MBP. A general version of the hunter and prey or "Pursuit" domain (Benda et al., 1986) and a beam walk domain have also been tested by MBP. The generalization of the hunter and prey domain is not described in detail in (Cimatti et al., 1998a). Thus, we have not been able to make an *NADL* implementation of this domain for a meaningful comparison.

The problem in the beam walk domain is for an agent to walk from one end of a beam to the other without falling down. If the agent falls, it has to walk back to the end of the beam and try again. The finite state machine of the domain is shown in Figure 13. The edges denote the outcome of a walk action. When the agent is on the beam, the walk action can either move it one step further on the beam or make it fall to a location under the beam.

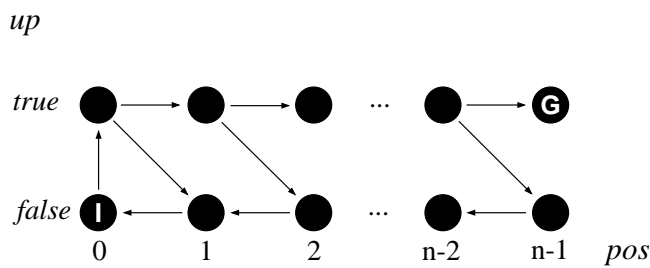

Figure 13: The beam walk domain. The *NADL* encoding of the beam walk domain has one propositional state variable *up* that is true if the agent is on the beam and false otherwise, and a numerical state variable *pos* that denotes the position of the agent either on the beam or on the ground. "I" and "G" are the initial state and goal state respectively.

We implemented a generator program for *NADL* descriptions of beam walk domains and produced domains with 4 to 4096 positions. Because the domain only contains two state variables, UMOP cannot exploit a partitioned transition relation for this domain, but has to use a monolithic representation.

The performance of UMOP and MBP is shown in Figure 14. Discounting that MBP was run on a slower machine,[5] the performance of UMOP and MBP is quite similar in this domain. For domains where UMOP can exploit a partitioned representation, we would expect it to be able to solve larger problems than MBP, since MBP currently can only use a monolithic representation. Further comparisons between UMOP and MBP are on our research agenda.

### 7.1.2 THE POWER PLANT DOMAIN

The purpose of the remaining experiments in non-deterministic domains is to show universal planning results for domains where the multi-agent and environment modelling features of *NADL* have been used.

The power plant domain demonstrates a multi-agent domain with an environment model and further exemplifies optimistic planning. It consists of reactors, heat exchangers, turbines and valves. A domain example is shown in Figure 15.

In the power plant domain each controllable unit is associated with an agent such that all control actions can be performed simultaneously. The environment consists of a single agent that at any time can fail a number of heat exchanges and turbines and also ensures that already failed units remain failed. A failed heat exchanger leaks water from the internal to the external water loop and must be closed by a block action *b*. The energy production from the reactor can be controlled by *p* to fit the demand *f*, but the reactor will always produce one energy unit. To transport the energy from the reactor away from the plant at least one heat exchanger and one turbine must be working. Otherwise the plant is in an unrecoverable failed state, where the reactor will overheat.

---

5. A 266MHz Pentium II with 96 MBytes RAM was used to generate the results for MBP.

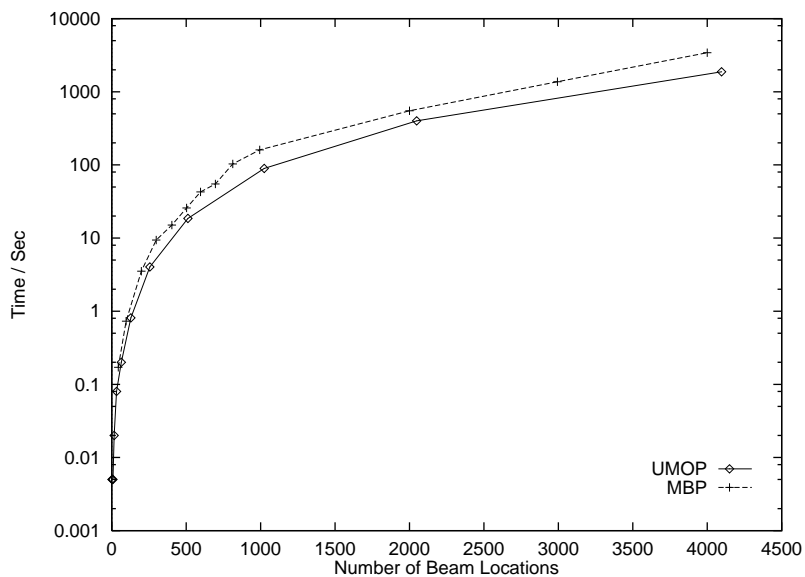

Figure 14: Planning time of UMOP and MBP in the beam walk domain. The MBP data has been extracted with possible loss of accuracy from (Cimatti et al., 1998a).

The state space of the power plant can be divided into three disjoint sets: good, bad and failed states. In the good states, therefore the goal states, the power plant satisfies its safety and activity requirements. In our example the safety requirements ensure that energy can be transported away from the plant, and that failed units are shut down:

```
% energy can be transported away from the plant
(okh1 \/ okh2 \/ okh3 \/ okh4) /\
(okt1 \/ okt2 \/ okt3 \/ okt4) /\

% heat exchangers blocked if failed
(~okh1 => b1) /\
(~okh2 => b2) /\
(~okh3 => b3) /\
(~okh4 => b4) /\

% turbines stopped if failed
(~okt1 => s1) /\
(~okt2 => s2) /\
(~okt3 => s3) /\
(~okt4 => s4)
```

The activity requirements state that the energy production equals the demand and that all valves to working turbines are open:

```
% power production equals demand
p = f /\

% turbine valve is open if turbine is ok
(okt1 => v1) /\
(okt2 => v2) /\
(okt3 => v3) /\
(okt4 => v4)
```

In a bad state, the plant does not satisfy the safety and activity requirements but is not unrecoverably failed. In a failed state all heat exchangers or turbines are failed.

The universal planning task is to generate a universal plan to get from any bad state to some good state without ending in a failed state. Assuming that no units fail during execution, it is obvious that only one joint action is needed. Unfortunately, the environment can fail any number of units during execution, thus, as described in Section 6.2, for any bad state the resulting joint action may loop back to a bad state or cause the plant to end in a failed state (see Figure 9). For this reason no strong or strong cyclic solution exist.

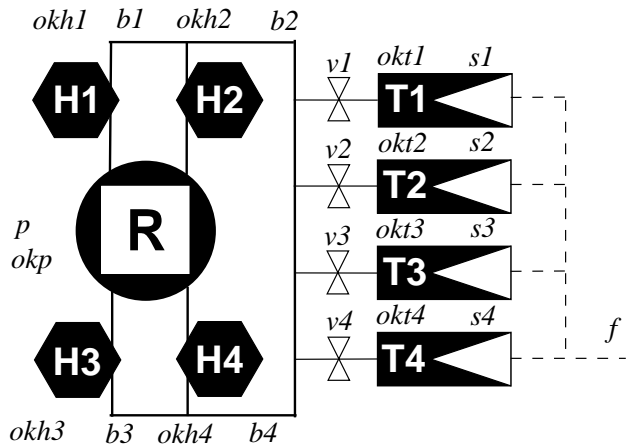

Figure 15: A power plant domain example. The reactor R is surrounded by four heat exchangers H1, H2, H3 and H4. The heat exchangers produce high pressure steam to four electricity generating turbines T1, T2, T3 and T4. A failed heat exchanger Hi must be closed by a block action $bi$. For a failed turbine Ti the stop action $si$ must be carried out. The energy production of the reactor is $p$ and can be controlled to fit the demand $f$. Each turbine Ti can be closed by a valve $vi$. The $ok$ variables capture the working status of the units.

An optimistic solution simply ignores that joint actions can loop back to a bad state or lead to a failed state and finds a solution to the problem after one preimage calculation. Intuitively, the optimistic plan assumes that no units will fail during execution and always chooses joint actions that lead directly from a bad state to a good state. The optimistic plan is an optimal control strategy, because it always chooses the shortest plan to a good state and no other strategy exists that can avoid looping back to a bad state or end in a failed state.

The size of the state space of the above power plant domain is $2^{24}$. An optimistic solution was generated by UMOP in 0.92 seconds and contained 37619 OBDD nodes. As an example, a joint action was extracted from the plan for a bad state where H3 and H4 were failed and energy demand $f$ was 2 energy units, while the energy production $p$ was only 1 unit. The extraction time was 0.013 seconds and, as expected, the set of joint actions included a single joint action changing $b3$ and $b4$ to true and setting $p$ to 2.

### 7.1.3 THE SOCCER DOMAIN

The purpose of the soccer domain is to demonstrate a multi-agent domain with a more elaborate environment model than the power plant domain. It consists of two teams of players that can move in a grid world and pass a ball to each other. At each time step a player either moves in one of the four major directions or passes the ball to another team player. The task is to generate a universal plan for one of the teams that can be applied to score a goal whenever the team possesses the ball.

A simple *NADL* description of the soccer domain models the team possessing the ball as system agents that can move and pass the ball independent of each other. Thus, a player possessing the ball can always pass to any other team player. The opponent team is modelled as a set of environment agents that can move in the four major directions but have no actions for handling the ball. The goal of the universal plan is to have a player with the ball in front of the opponent goal without having any opponents in the goal area.

It is impossible to generate a strong plan that covers all possible initial states. For instance an initial state with an opponent located in the goal area has no strong solution. But a strong plan covering as many initial states as possible is still useful, because it defines all the "scoring" states of the game and further provides a plan for scoring the goal no matter the actions, the opponent players choose.

We implemented an *NADL* generator for soccer domains with different field sizes and numbers of agents. The multi-agent graph in Figure 16 shows UMOP's planning time using the strong planning algorithm in soccer domains with 64 locations and one to six players on each team. The planning time seems to grow exponentially with the number of players. This is not surprising as not only the state space but also the number of joint actions grow exponentially with the number of agents. To investigate the complexity introduced by joint actions we constructed a version of the soccer domain with only a single system and environment agent and ran UMOP again. The single-agent graph in Figure 16 shows the dramatic decrease in computation time. It is not obvious though, that using more agents increases the computational load, as this normally also reduces the number of preimage calculations, because a larger number of states is reached in each iteration. Indeed, in a version of the power plant domain with deterministic actions, we found the planning time to decrease (see the power plant graph in Figure 16), when more agents were added (Jensen, 1999). Again we measured the time for extracting actions from the generated universal plans. For the multi-agent version of the five player soccer domain the two joint actions achieving the goal shown in Figure 17 were extracted from the universal plan in less than 0.001 seconds.

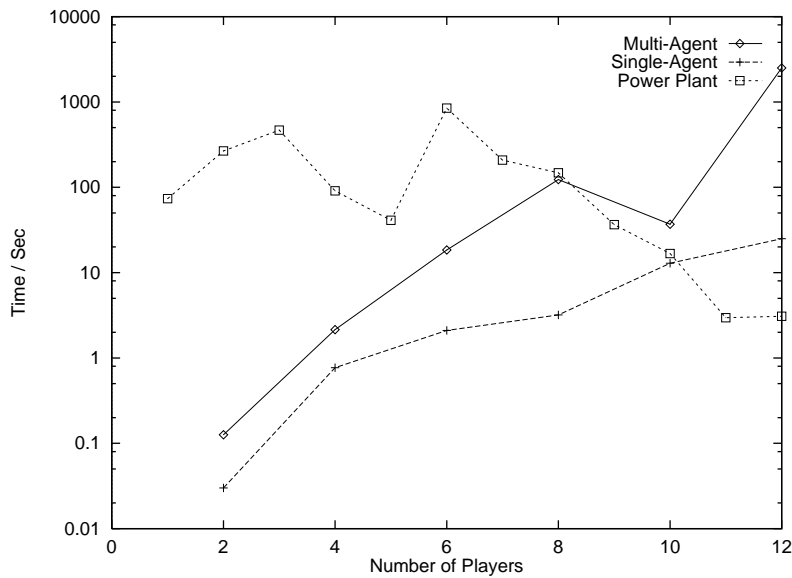

Figure 16: Planning time of UMOP for generating strong universal plans in soccer domains with one to six players on each team. For the multi-agent experiment each player was associated with an agent, while only a single system and environment agent was used in the single-agent experiment. The power plant graph shows planning time for a deterministic version of the power plant domain using 1 to 12 system agents.

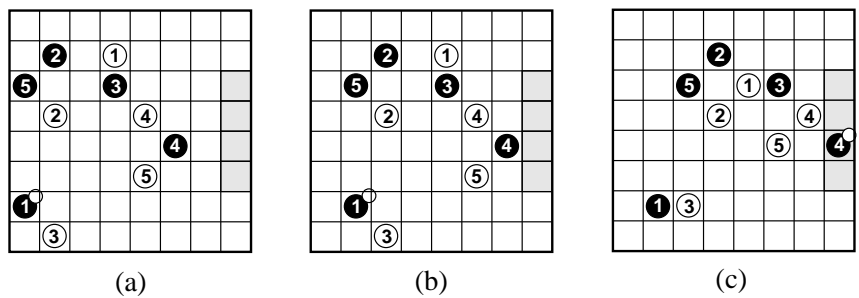

(a)                    (b)                    (c)

Figure 17: Plan execution sequence. The three states show a hypothetical attack based on a universal plan. The state (a) is a "scoring" state, because the attackers (black) can extract a nonempty set of joint actions from the universal plan. Choosing some joint actions from the plan, the attackers can enter the goal area (shaded) with the ball within two time steps (state (b) and (c)) no matter what actions, the opponent players choose.

## 7.2 Deterministic Domains

A number of experiments have been carried out in deterministic domains in order to verify UMOP's performance and illustrate the generality of universal plans versus classical, se-quential plans. We compare run time results obtained with UMOP in some of the AIPS'98 competition domains to the results of the competition planners (McDermott, 1999). We then show that a universal plan in a deterministic domain is more general than a classical sequential plan, because a large number of classical sequential plans are contained in the universal plan.

### 7.2.1 AIPS'98 COMPETITION DOMAINS

Four planning systems BLACKBOX (Kautz & Selman, 1999), IPP (Koehler et al., 1997), STAN (Long & Fox, 1998) and HSP (Haslum & Geffner, 2000)[6] competed in the three domains we have studied. BLACKBOX is based on SATPLAN, while IPP and STAN are graphplan-based planners. HSP uses a heuristic search approach based on a preprocessing of the domain. The AIPS'98 planners were run on 233 MHz (or 400MHz)[7] Pentium PCs with 128 MBytes RAM equipped with Linux.

**The Gripper Domain.** The gripper domain consists of two rooms A and B, a robot with a left and right gripper and a number of balls that can be moved by the robot. The task is to move all the balls from room A to room B with the robot initially in room A. The state variables of the *NADL* encoding of the domain are the position of the robot and the position of the balls. The position of the robot is either 0 (room A) or 1 (room B), while the position of a ball can be 0 (room A), 1 (room B), 2 (in left gripper) or 3 (in right gripper). For the AIPS'98 gripper problems the number of plan steps in an optimal plan grows linearly with the problem number. Problem 1 contains 4 balls, and the number of balls grows by two for each problem. The result of the experiment is shown in Table 2 together with the results of the planners in the AIPS'98 competition. A graphical representation of the planning time in the table is shown in Figure 18. UMOP generates minimum-length plans due to its parallel breadth-first search algorithm. As depicted in Figure 18, it avoids the exponential growth of the planning time that characterizes all of the competition planners except HSP. When using a partitioned transition relation UMOP is the only planner capable of generating optimal plans for all the problems. For this domain the transition relation of an *NADL* description can be divided into $n + 1$ basic partitions, where $n$ is the number of balls. As discussed in Section 5, the optimal number of partitions is not necessarily the largest number of partitions. For the results in Table 2 each partition consisted of a conjunction of 10 basic partitions. Compared to the monolithic transition relation representation the results obtained with the partitioned transition relation were significantly better on the larger problems. The memory usage of problem 20 with a partitioned transition relation was 87 MBytes, while it exceeded the limit of 128 MBytes at problem 17 for the monolithic transition relation.

**The Movie Domain.** In the movie domain the task is to get chips, dip, pop, cheese and crackers, rewind a movie and set the counter to zero. The only interference between the subgoals is that the movie must be rewound, before the counter can be set to zero. The problems in the movie domain only differ by the number of objects of each type of food. The number of objects increases linearly from 5 for Problem 1 to 34 for Problem 30.

---

6. PRODIGY4.0 also successfully ran in some of the domains, but was not an official entry in the competition.

7. Unfortunately, no exact record has been kept on the machines and there is some disagreement about their clock frequency. According to Drew McDermott, who chaired the competition, they were 233 MHz Pentiums, but Derek Long (STAN) believes that they were at least 400 MHz Pentiums, as STAN performed worse on a 300 MHz Pentium than in the competition.

| Problem | UMOP Part. | | | UMOP Mono. | | STAN | | HSP | | IPP | | BLACKBOX | |
|---|---|---|---|---|---|---|---|---|---|---|---|---|---|
| 1 | 20 | 11 | 1 | 20 | 11 | 46 | 11 | 2007 | 13 | 50 | 15 | 113 | 11 |
| 2 | 150 | 17 | 1 | 130 | 17 | 1075 | 17 | 2150 | 21 | 380 | 23 | 7820 | 17 |
| 3 | 710 | 23 | 1 | 740 | 23 | 54693 | 23 | 2485 | 31 | 3270 | 31 | - | - |
| 4 | 1490 | 29 | 2 | 2230 | 29 | 3038381 | 29 | 3060 | 37 | 26680 | 39 | - | - |
| 5 | 3600 | 35 | 2 | 6040 | 35 | - | - | 3320 | 47 | 226460 | 47 | - | - |
| 6 | 7260 | 41 | 2 | 11840 | 41 | - | - | 3779 | 53 | - | - | - | - |
| 7 | 13750 | 47 | 2 | 24380 | 47 | - | - | 4797 | 63 | - | - | - | - |
| 8 | 23840 | 53 | 2 | 38400 | 53 | - | - | 5565 | 71 | - | - | - | - |
| 9 | 36220 | 59 | 3 | 68750 | 59 | - | - | 6675 | 79 | - | - | - | - |
| 10 | 56200 | 65 | 3 | 95140 | 65 | - | - | 7583 | 85 | - | - | - | - |
| 11 | 84930 | 71 | 3 | 145770 | 71 | - | - | 9060 | 93 | - | - | - | - |
| 12 | 127870 | 77 | 3 | 216110 | 77 | - | - | 10617 | 101 | - | - | - | - |
| 13 | 197170 | 83 | 3 | 315150 | 83 | - | - | 12499 | 109 | - | - | - | - |
| 14 | 290620 | 89 | 4 | 474560 | 89 | - | - | 15050 | 119 | - | - | - | - |
| 15 | 411720 | 95 | 4 | 668920 | 95 | - | - | 16886 | 125 | - | - | - | - |
| 16 | 549610 | 101 | 4 | 976690 | 101 | - | - | 20084 | 135 | - | - | - | - |
| 17 | 746920 | 107 | 4 | - | - | - | - | 23613 | 143 | - | - | - | - |
| 18 | 971420 | 113 | 4 | - | - | - | - | 26973 | 151 | - | - | - | - |
| 19 | 1361580 | 119 | 5 | - | - | - | - | 29851 | 157 | - | - | - | - |
| 20 | 1838110 | 125 | 5 | - | - | - | - | 33210 | 165 | - | - | - | - |

Table 2: Gripper domain results. Column one and two for each planner show the planning time in milliseconds and the plan length. UMOP Part. and UMOP Mono. show the planning time for UMOP using a partitioned and a monolithic transition relation, respectively. For UMOP with partitioned transition relation the third column shows the number of partitions. (-) means that the planner used more than 128 MBytes of memory or was terminated before returning a solution. Only results for executions using less than 128 MBytes are shown for UMOP.

Our *NADL* description of the movie domain represents each type of food as a numerical state variable with a range equal to the number of objects of that type of food. Table 3 shows the planning time for UMOP and the competition planners for the movie domain problems. In this experiment and the remaining experiments UMOP used its default partitioning of the transition relation. For every problem all the planners find the optimal solution. Like most of the competition planners UMOP has a low computation time, but it is the only planner not showing any increase in computation time even though, the size of the state space of its encoding increases from $2^{24}$ to $2^{39}$.

**The Logistics Domain**. The logistics domain (Veloso, 1994) consists of cities, trucks, airplanes and packages. Trucks can only move between locations in the same city. Airplanes can only move between airport locations in different cities. The task is to move packages to specific locations. Problems differ by the number of packages, cities, airplanes and trucks. The logistics domain is hard, and only Problem 1,2,5,7 and 11 of the 30 problems were solved by any planner in the AIPS'98 competition (see Table 4). The *NADL* description of the logistics domain uses numerical state variables to represent locations of packages, where trucks and airplanes are treated as special locations. Even though the state space of the small problems is moderate, UMOP fails to solve any of the problems in the domain. It succeeds to generate the transition relation but fails to finish the preimage calculations.

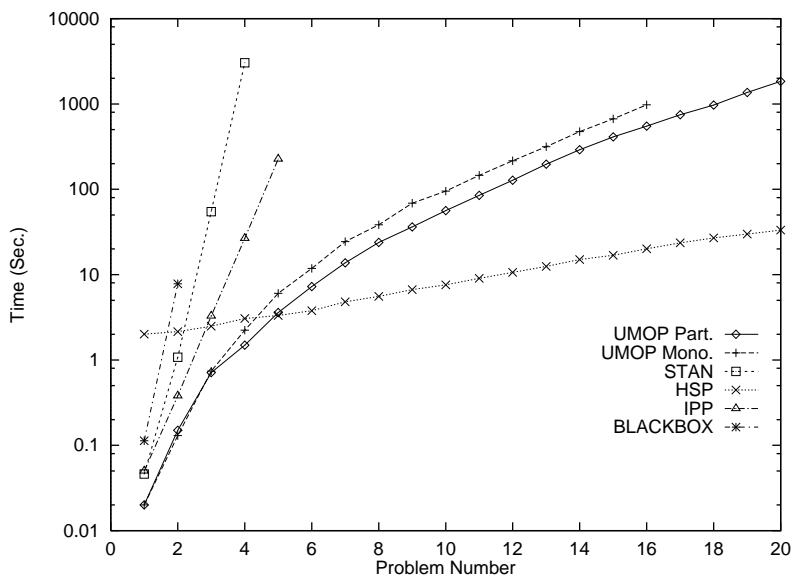

Figure 18: Planning time for UMOP and the AIPS'98 competition planners for the gripper domain problems. UMOP Part. and UMOP Mono. show the planning time for UMOP using a partitioned and a monolithic transition relation, respectively.

We have studied the logistics domain extensively, recently focusing on OBDD-based deterministic planning. The logistics domain seems to be hard using a plain OBDD-based approach, as the sizes of the preimages grow too fast. To address this complexity, we have developed an abstraction technique for OBDD-based deterministic planning. In a nutshell, a problem is first solved using an abstract transition system, where each transition corresponds to a set of serializable actions. Then the steps in the abstract plan are serialized using an ordinary transition system. With this new algorithm, we have been able to solve several of the complex AIPS'98 competition logistics problems (Jensen et al., 2000).

### 7.2.2 THE OBSTACLE DOMAIN

The obstacle domain has been constructed to demonstrate the generality of universal plans. It consists of a grid world with $2^5$ cells, $n$ obstacles and a robot agent. The positions of the obstacles are not defined. The goal position of the robot is the upper right corner of the grid, and the task for the robot is to move from any position in the grid to the goal position. Because the initial locations of obstacles are unknown, the universal plan must take any possible position of obstacles into account, which gives $2^{5(n+1)} - 2^{5n}$ initial states. For a specific initial state a sequential plan can be generated from the universal plan. Thus, $2^{5(n+1)} - 2^{5n}$ sequential plans are comprised in one universal plan. Note that a universal plan with $n$ obstacles includes any universal plan with 1 to $n$ obstacles, as obstacles can be placed at the same location. Note moreover, that the universal plans never cover all initial states, because obstacles can be placed at the goal position, and obstacles can block the robot.

| Problem | UMOP | STAN | HSP | IPP | BLACKBOX |
|---------|------|------|------|-----|----------|
| 1 | 14 | 19 | 2121 | 10 | 11 |
| 2 | 12 | 18 | 2104 | 10 | 12 |
| 3 | 14 | 19 | 2144 | 10 | 14 |
| 4 | 4 | 20 | 2188 | 10 | 16 |
| 5 | 14 | 21 | 2208 | 10 | 18 |
| 6 | 16 | 22 | 2617 | 10 | 20 |
| 7 | 14 | 22 | 2316 | 20 | 22 |
| 8 | 12 | 23 | 2315 | 20 | 24 |
| 9 | 14 | 25 | 2357 | - | 26 |
| 10 | 14 | 26 | 2511 | 10 | 29 |
| 11 | 14 | 27 | 2427 | 30 | 30 |
| 12 | 4 | 28 | 2456 | 30 | 32 |
| 13 | 16 | 29 | 3070 | 20 | 36 |
| 14 | 14 | 31 | 2573 | 30 | 35 |
| 15 | 16 | 32 | 2577 | 30 | 38 |
| 16 | 14 | 34 | 2699 | 10 | 39 |
| 17 | 16 | 35 | 2645 | 30 | 41 |
| 18 | 14 | 37 | 2686 | 10 | 43 |
| 19 | 16 | 39 | 2727 | 30 | 45 |
| 20 | 12 | 40 | 2787 | 20 | 47 |
| 21 | 16 | 42 | 2834 | 20 | 49 |
| 22 | 14 | 45 | 2834 | 20 | 51 |
| 23 | 16 | 48 | 2866 | 20 | 53 |
| 24 | 14 | 50 | 3341 | 20 | 55 |
| 25 | 16 | 52 | 2997 | 30 | 57 |
| 26 | 16 | 54 | 3013 | 40 | 58 |
| 27 | 16 | 57 | 3253 | 50 | 60 |
| 28 | 4 | 62 | 3049 | 40 | 63 |
| 29 | 18 | 64 | 3384 | 50 | 64 |
| 30 | 16 | 67 | 3127 | 40 | 66 |

Table 3: Movie domain results. The table shows the run time in milliseconds for each planner. (-) means that the planner used more than 128 MBytes of memory or was terminated before returning a solution. All planners generated optimal plans of length 7. UMOP used far less than 128 MBytes for any problem in this domain.

A universal plan for an obstacle domain with 5 obstacles was generated with UMOP in 420 seconds and contained 488296 OBDD nodes (13.3 MBytes). Sequential plans were extracted from the universal plan for a specific position of the obstacles. Figure 19 shows the extraction time of sequential plans for an increasing number of steps in the plan. Even though the OBDD representing the universal plan is large, the extraction is very fast and only grows linearly with the plan length.

The set of actions associated with a state $s$ in a universal plan $p$ is extracted by computing the conjunction of the OBDD representation of $s$ and $p$. As described in Section 3, this operation has an upper bound complexity of $O(|s||p|)$. For the universal plan in the obstacle domain with five obstacles, this computation was fast (less than one millisecond) and would allow an executing robot to meet low reaction time constraints.

| Problem | UMOP | | STAN | | HSP | | IPP | | BLACKBOX | |
|---------|------|---|--------|----|--------|-----|------|----|----------|----|
| 1 | - | - | 767 | 27 | 79682 | 43 | 900 | 26 | 2062 | 27 |
| 2 | - | - | 4319 | 32 | 97114 | 44 | - | - | 6436 | 32 |
| 5 | - | - | 364932 | 29 | 144413 | 26 | 2400 | 24 | - | - |
| 7 | - | - | - | - | 788914 | 112 | - | - | - | - |
| 11 | - | - | 12806 | 34 | 86195 | 30 | 6940 | 33 | 6544 | 32 |

Table 4: Logistics domain results. For each planner column one and two show the run time in milliseconds and the plan length. (-) means that the planner used more than 128 MBytes of memory or was terminated before returning a solution.

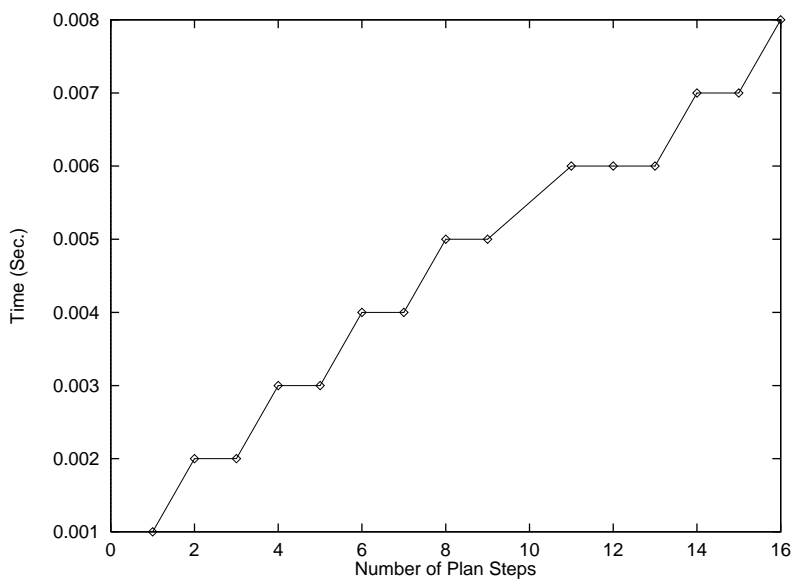

Figure 19: Time for extracting sequential plans from a universal plan for the obstacle domain with 5 obstacles.

## 8. Conclusion and Future Work

In this article we have presented a new OBDD-based planning system, UMOP, for planning in non-deterministic, multi-agent domains. An expressive domain description language, *NADL*, has been developed and an efficient OBDD representation of its NFA semantics has been described. We have analyzed previous planning algorithms for OBDD-based planning and deepened the understanding of when these planning algorithms are appropriate. Finally, we have proposed an optimistic planning algorithm for finding sensible solutions in some domains where no strong or strong cyclic solution exists. The results obtained with UMOP are encouraging, as UMOP has a good performance compared to some of the fastest classical planners known today.

Our research has drawn our attention to a number of open questions that we would like to address in the future. In particular we wonder how well our encoding of planning problems scales compared to the encoding used by MBP. Currently MBP's encoding does not support a partitioned representation of the transition relation, but the encoding may have other properties that, despite the monolithic representation, may make it a better choice. The two systems may also have an equal performance when both are using a monolithic representation (as in the beam walk example), which should give UMOP an advantage in domains where a partitioning of the transition relation can be defined.

Another interesting question is to investigate which kind of planning domains is suitable for OBDD-based planning. It was surprising for us that the logistics domain turned out to be so hard for UMOP. Recently we have studied this domain thoroughly. Using an abstraction technique we have now been able to solve several of the logistics problems in the AIPS'98 competition (Jensen et al., 2000).

The current definition of *NADL* is powerful but should be extended to enable modelling of constructive synergetic effects as described in Section 4. Also, we envision more experiments comparing multi-agent and single-agent domains to investigate the complexity of *NADL*'s representation of concurrent actions.

Several planners, in particular PRODIGY (Veloso et al., 1995), have shown that domain knowledge should be used by a planning system in order to scale up to real-world problems. Also (Bacchus & Kabanza, 1996) show how the search tree of a forward chaining planner can be efficiently pruned by stating the goal as a formula in temporal logic on the sequence of actions leading to the goal. In this way the goal can include knowledge about the domain (e.g., that towers in the blocks world must be built from bottom to top). A similar approach for reducing the complexity of OBDD-based planning seems promising, especially because techniques for testing temporal formulas already have been developed in model checking.

Other future challenges include introducing abstraction in OBDD-based planning and defining specialized planning algorithms for multi-agent domains (e.g., algorithms using the least number of agents for solving a problem).

## Acknowledgments

Special thanks to Paolo Traverso, Marco Roveri and the other members of the IRST group for introducing us to MBP and for many rewarding discussions on OBDD-based planning and model checking. We also wish to thank Randal E. Bryant, Edmund Clarke, Henrik R. Andersen, Jørn Lind-Nielsen and Lars Birkedal for advice on OBDD issues and formal representation. Finally, we thank the anonymous reviewers for their comments that greatly improved the presentation of this article.

This work was carried out while the first author was visiting Carnegie Mellon University from the Technical University of Denmark. The research is sponsored in part by McKinsey & Company, Selmer & Trane's Fond, the Defense Advanced Research Projects Agency (DARPA) and the Air Force Research Laboratory (AFRL) under agreement number F30602-97-2-0250. The views and conclusions contained herein are those of the authors and should not be interpreted as necessarily representing the official policies or endorsements, either expressed or implied, of the Defense Advanced Research Projects Agency (DARPA), the Air Force Research Laboratory (AFRL) or the U.S. Government.

## Appendix A. *NADL* includes the $\mathcal{AR}$ Family

**Theorem 1** *If $A$ is a domain description in some $\mathcal{AR}$ language, then there exists a domain description $D$ in NADL with the same semantics as $A$.*

*Proof*: Let $M_a = (Q, \Sigma, \delta)$ denote the NFA (see Definition 1) corresponding to the semantics of $A$ as defined by Giunchiglia et al. (1997). An *NADL* domain description $D$ with semantics equal to $M_a$ can be constructed in the following way: Let $D$ be a single-agent domain where all fluents are encoded as numerical state variables and there is an action for each element in the alphabet $\Sigma$ of $M_a$. Consider the action $a$ associated to input $i \in \Sigma$. Let the set of constrained state variables of $a$ equal the set of state variables in $D$. The precondition of $a$ is an expression that defines the set of states having an outgoing transition for input $i$. The effect condition of $a$ is a conjunction of conditional effects ($P_s \Rightarrow N_s$). There is one conditional effect for each state that has an outgoing transition for input $i$. $P_s$ in the conditional effect associated with state $s$ is the characteristic expression for $s$ and $N_s$ is a characteristic expression for the set of next states $\delta(s, i)$. $\square$